\documentclass[conference]{IEEEtran}
\DeclareUnicodeCharacter{2212}{-}
\usepackage{cite}
\usepackage{url}
\usepackage{amsmath,amssymb,amsfonts}
\usepackage{algorithmic}
\usepackage{graphicx}
\usepackage{textcomp}
\usepackage{xcolor}
\usepackage{multirow}
\usepackage{tabularx}
\usepackage{supertabular}
\usepackage{subcaption}
\usepackage{eurosym}
\newcount\myloopcounter
\newcommand{\nstars}[1][4]{%
  \myloopcounter0% initialize the loop counter
  \loop\ifnum\myloopcounter < 4
  \ifthenelse{\myloopcounter < #1}{
    \textcolor{black}{\star}
  }{
    \textcolor{black!22}{\star}
  }
  \advance\myloopcounter by 1 % 
  \repeat % start again
}

\usepackage[draft=False]{hyperref} % put False when you've finished all your writing
\hypersetup{
     colorlinks=true,
     linkcolor=blue,
     filecolor=blue,
     citecolor=blue,      
     urlcolor=blue,
     }
\def\BibTeX{{\rm B\kern-.05em{\sc i\kern-.025em b}\kern-.08em
    T\kern-.1667em\lower.7ex\hbox{E}\kern-.125emX}}
\begin{document}

\title{Denoising diffusion probabilistic models for probabilistic energy forecasting}

\author{\IEEEauthorblockN{Esteban Hernandez Capel}
\IEEEauthorblockA{\textit{Smart-Microgrids department} \\
\textit{Uliege}\\
Liege, Belgium \\
hernandez.capel.esteban@gmail.com}
\and
\IEEEauthorblockN{Jonathan Dumas}
\IEEEauthorblockA{\textit{R\&D department} \\
\textit{Réseau de Transport d’Électricité (RTE)}\\
Paris, France \\
jonathan.dumas@rte-france.com}
}

\maketitle

\begin{abstract}
Scenario-based probabilistic forecasts have become vital for decision-makers to handle intermittent renewable energies.
This paper presents a recent promising deep learning generative approach called \textit{denoising diffusion probabilistic models}. It is a class of latent variable models which have recently demonstrated impressive results in the computer vision community. 
However, to our knowledge, there has yet to be a demonstration that they can generate high-quality samples of load, PV, or wind power time series, crucial elements to face the new challenges in power systems applications.
Thus, we propose the first implementation of this model for energy forecasting using the open data of the Global Energy Forecasting Competition 2014. The results demonstrate this approach is competitive with other state-of-the-art deep learning generative models, including generative adversarial networks, variational autoencoders, and normalizing flows. 
\end{abstract}


\section{Introduction}

The Intergovernmental Panel on Climate Change in the AR6 report \cite{ipccar6} set ambitious targets to decrease greenhouse gas emissions, and the energy transition requires a necessary growth of renewable generation in the energy mix.
However, renewable energies \textit{e.g.}, solar and wind power, are subject to uncertainty in contrast to conventional power plants. Thus, they have challenged the operational predictability of modern power systems. 
In this context, decision-makers have used \textit{probabilistic} forecasts as an essential tool to improve decisions in various applications of power systems \cite{morales2013integrating}.

% Different types of forecasts
Power systems forecasting practitioners employ various types of probabilistic forecasting techniques ranging from \textit{quantile} to \textit{density forecasts}, \textit{scenarios}, and through \textit{prediction intervals} \cite{morales2013integrating}.
This paper concentrates on scenario generation by using \textit{deep-learning generative} approaches to model time series of load, photovoltaic (PV), and wind power generations. They train deep neural networks to model the distribution of these random variables. 
Various deep-learning generative approaches have been studied in the literature, such as variational autoencoders (VAEs), generative adversarial networks (GANs), normalizing flows (NFs), and numerous hybrid strategies. Each technique has pros and cons, and the selection depends on trade-offs regarding computation time, quality and value results, and architectural restrictions. 
We recommend three papers to get a broader knowledge of this field.
(1) The comprehensive overview of generative modeling trends conducted by \cite{bond2021deep}. It presents generative models to forecasting practitioners under a single cohesive statistical framework. (2) The thorough comparison of normalizing flows, variational autoencoders, and generative adversarial networks provided by \cite{ruthotto2021introduction}. It describes the advantages and disadvantages of each approach using numerical experiments in computer vision. (3) Finally, \cite{DUMAS2022117871} propose applying conditional generative models in power systems.

This paper proposes investigating a new promising deep generative modeling method for energy forecasting: \textit{denoising diffusion probabilistic models} (DDPMs). 
DDPMs are a class of likelihood-based models that have recently demonstrated remarkable results in computer vision \textit{e.g.}, \cite{DDPM} and \cite{diff_beat_gan}, and the natural language processing communities \textit{e.g.}, \cite{diff_wave} and \cite{Diff_TTS}. 
A DDPM is a parameterized Markov chain trained using variational inference to produce samples matching the data after a finite time. Transitions of this chain are learned to reverse a diffusion process. This Markov chain gradually adds noise to the data in the opposite direction of sampling until the signal is destroyed. 
They offer desirable properties such as distribution coverage, a stationary training objective, and manageable scalability. These models generate samples by gradually removing noise from a signal, and their training objective is expressed as a reweighted variational lower-bound \cite{DDPM}.
However, they have not been implemented and tested in power system applications and rarely in time series settings \cite{DDPM_forecast}. 

This study is an extension of \cite{DUMAS2022117871} based on \cite{capeldenoising} to bridge this research gap with two main contributions.
First, we use the open data of the Global Energy Forecasting Competition 2014 (GEFcom 2014) \cite{hong2016bprobabilistic} to compare, in terms of quality and value, DDPMs with state-of-the-art deep learning generative models: NFs, GANs, and VAEs. To our knowledge, this study is the first to i) implement a DDPM in an energy forecasting application; ii) compare with complementary metrics on an easily reproducible case study DDPMs to NFs, GANs, and VAEs on several datasets, including PV and wind generation, and load.
Second, in this case study, DDPMs achieved better results in quality and value. It provides evidence for deep learning practitioners further to study this model in more advanced power system applications.
In addition, this study provides open access to the Python code\footnote{\url{https://github.com/EstebanHernandezCapel/DDPM-Power-systems-forecasting}} to help the community to reproduce the experiments.

\begin{figure}[tb]
	\centering
\includegraphics[width=\linewidth]{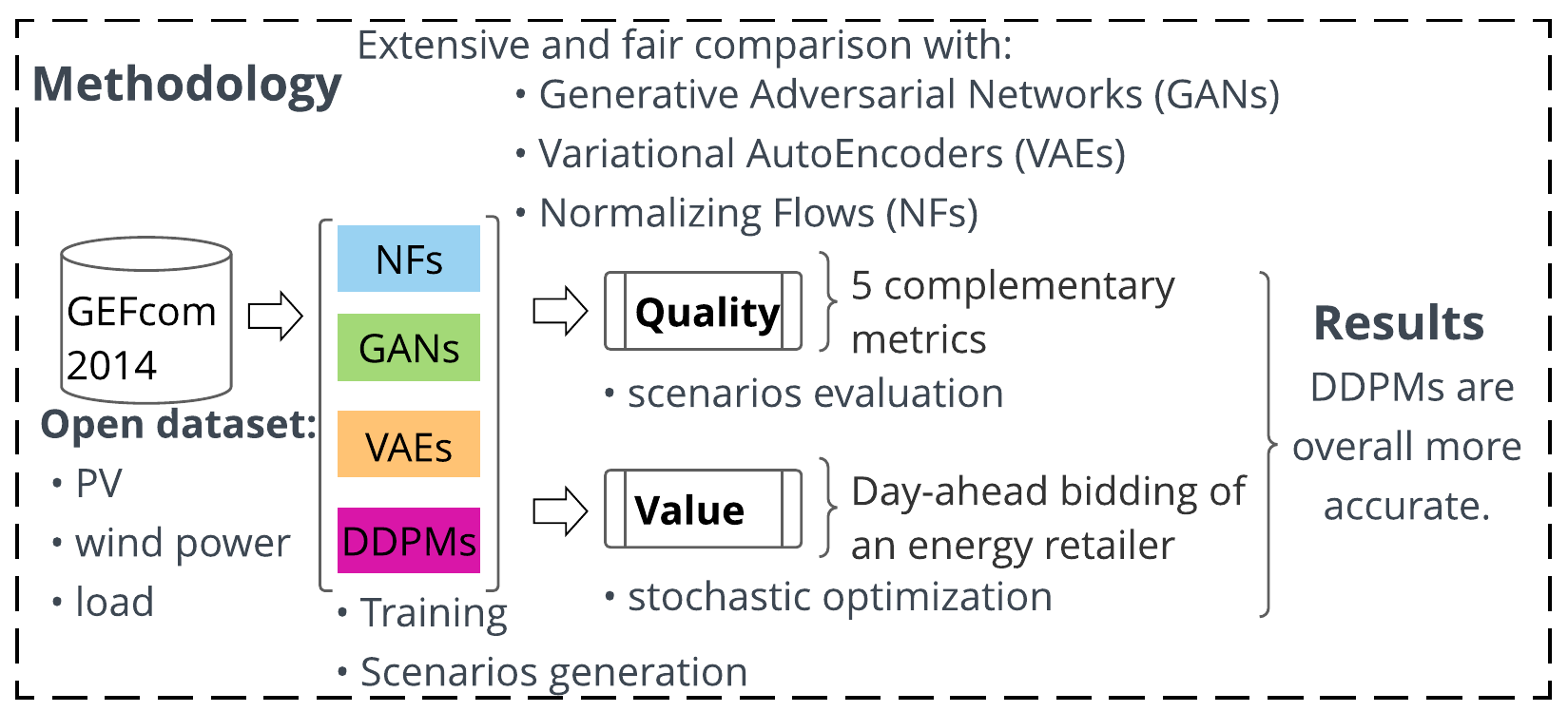}
\caption{The framework of the paper.}
\label{fig:paper-framework}
\end{figure}

Fig.~\ref{fig:paper-framework} depicts the framework of the proposed method, and the remainder of this paper is organized as follows. 
Section~\ref{sec:background} presents the four generative models implemented. Section~\ref{sec:quality_assessment} provides the quality and value assessment methodologies. Section~\ref{sec:numerical_results} details empirical results on the GEFcom 2014 dataset, and Section~\ref{sec:conclusion} summarizes the main findings and highlights ideas for further work. 

\section{Background}\label{sec:background}

This section provides a high-level description of DDPMs and a reminder of the basic principles of the conditional version of GANs, VAEs, and NFs implemented in \cite{DUMAS2022117871}, used to challenge DDPMs. 

\subsection{Daily scenario generation}

We consider a dataset $\mathcal{D} = \{ \mathbf{x}^i, \mathbf{c}^i \}_{i=1}^N$ composed of $N$ independent and identically distributed samples from the joint distribution $p(\mathbf{x},\mathbf{c})$ of two continuous variables $X$ and $C$. 
$X$ is a time series such as load, wind, or PV generation, and $C$ is composed of weather forecasts that provide relevant information such as wind speed, solar irradiation, or temperature.
They are both composed of $T$ periods per day, with $\mathbf{x}^i := [x_1^i, \ldots , x_T^i]^\intercal \in \mathbb{R}^T$ and $\mathbf{c}^i := [c_1^i, \ldots , c_T^i]^\intercal \in \mathbb{R}^T$. 
This paper proposes to generate weather-based scenarios $\mathbf{\hat{x}} \in \mathbb{R}^T$ that are distributed under $p(\mathbf{x}|\mathbf{c})$ by using a deep-learning generative model.
It is a probabilistic model $p_\theta(\cdot)$, with parameters $\theta$, used to generate synthetic but realistic data $\mathbf{\hat{x}} \sim p_\theta(\mathbf{x}|\mathbf{c})$ whose distribution is as close as possible to the PV, wind power, or load, unknown distribution $p(\mathbf{x}|\mathbf{c})$. 
In this paper, the generative model computes a set of $M$ scenarios at day $d-1$ for each day $d$ of the dataset
\begin{align}
	\label{eq:multi_output_scenario}	
	\mathbf{\hat{x}}_d^i := &  \big[\hat{x}_{d, 1}^i, \cdots,\hat{x}_{d, T}^i\big]^\intercal	\in \mathbb{R}^T \quad i=1, \ldots, M.
\end{align}
For clarity, in the following, we omit the indexes $d$ and $i$ when referring to a scenario $\mathbf{\hat{x}}$. 

\subsection{High-level description of deep generative models}

\begin{figure}[tb]
	\centering
\includegraphics[width=\linewidth]{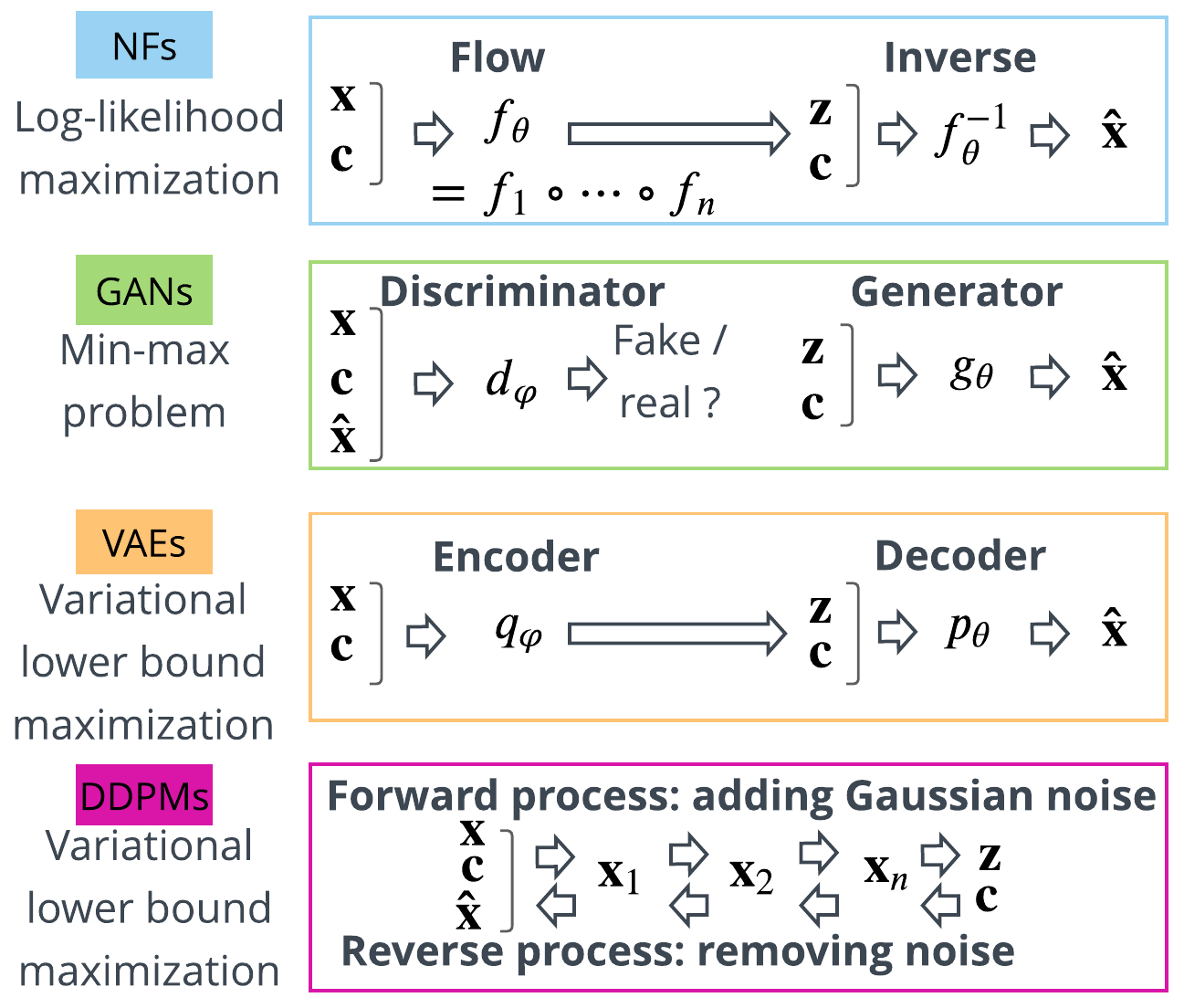}
\caption{High-level comparison of the generative models: normalizing flows, generative adversarial networks, variational autoencoders, and denoising diffusion probabilistic models.}
\label{fig:methods-comparison}
\end{figure}

Fig.~\ref{fig:methods-comparison} depicts a high-level description of the generative models considered.
All models are conditional as they use the weather forecasts $\mathbf{c}$ to generate scenarios $\mathbf{\hat{x}}$ of the distribution of interest $\mathbf{x}$: PV generation, wind power, load.
%
% NFs
NFs allow exact likelihood calculation. In contrast to GANs, VAEs, and DDPMs, they explicitly learn the data distribution and directly access the exact likelihood of the model's parameters. The inverse of the flow is used to generate scenarios.
% GANs 
The training of GANs relies on a min-max problem where the generator and the discriminator parameters are jointly optimized. The generator is used to compute the scenarios.
% VAEs
VAEs indirectly optimize the log-likelihood of the data by maximizing the variational lower bound. The decoder computes the scenarios.
% DDPMs
Finally, DDPMs are a class of likelihood-based models. They generate samples by gradually removing noise from a signal, and their training objective can be expressed as a reweighted variational lower bound. A strength of DDPMs is the loss function from the variational lower bound expressed in Kullback-Leibler to a straightforward MSE between the Gaussian noises.

\subsection{Denoising diffusion models}

\begin{figure}[bt]
	\centering
\includegraphics[width=\linewidth]{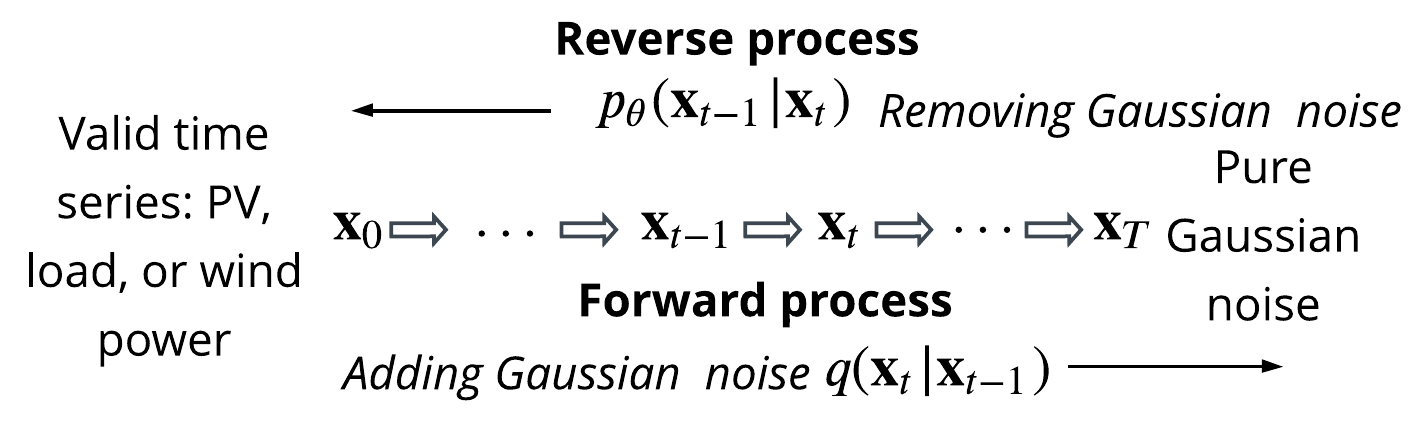}
\caption{Illustration of the DDPM forward and reverse processes. The forward process $q(\mathbf{x}_t|\mathbf{x}_{t-1})$ uses a data sample $\mathbf{x}_0$ and gradually adds Gaussian noise to produce noised samples $\mathbf{x}_1$ trough $\mathbf{x}_T$. The reverse process $p_{\theta}(\mathbf{x}_{t-1}|\mathbf{x}_t)$ gradually removes the noise to generate a realistic load scenario.}
    \label{fig:DDPM_illustration}
\end{figure}

This section provides the basic principles of the DDPM, and more details are provided in \cite{capeldenoising}.
In this section, the number of periods per day $T$ is denoted $L$ to avoid confusion with the number of diffusion steps noted $T$. For clarity, we omit the conditional vector $\mathbf{c}$, and a sample of the original distribution $\mathbf{x} \sim p(\mathbf{x},\mathbf{c})$ is noted $\mathbf{x}_0 \sim p(\mathbf{x}_0)$. 
On a high level, diffusion models sample from a distribution by reversing a gradual noising process. In particular, sampling starts with noise $\mathbf{x}_T$ and produces less-noisy samples gradually $\mathbf{x}_{T-1}$, $\mathbf{x}_{T-2}$, ..., until reaching a final sample $\mathbf{x}_0$. Each step $i$ corresponds to a certain noise level, and $\mathbf{x}_i$ can be interpreted as a combination of a signal $\mathbf{x}_0$ with some noise where the step $i$ determines the signal to noise ratio. The model implemented uses a noise drawn from a diagonal Gaussian distribution \cite{diff_beat_gan}.
Formally, DDPM is a class latent variable model of the form
\begin{align}
 p_{\theta}(\mathbf{x}_{0}):=\int p_{\theta}(\mathbf{x}_{0: T}) d \mathbf{x}_{1:T},
\end{align}
where $\mathbf{x}_1, \mathbf{x}_2,..., \mathbf{x}_T$ are latent variables of the same dimensionality as the data $\mathbf{x}_0 \sim p_\theta(\mathbf{x}_0)$, with $\mathbf{x}_0 \in \mathbb{R}^L$. Overall, a DDPM comprises two main parts depicted by Fig.~\ref{fig:DDPM_illustration}.

First, the joint distribution $p_{\theta}(\mathbf{x}_{0:T})$, named \textit{reverse process}, is defined by a Markov chain with Gaussian transitions starting at $p(\mathbf{x}_{T}) = \mathcal{N}(\mathbf{x}_T; \mathbf{0} ; \mathbf{I})$
\begin{align}
p_{\theta}(\mathbf{x}_{0:T}):= & p(\mathbf{x}_{T}) \prod_{i=1}^{T} p_{\theta}(\mathbf{x}_{i-1} \mid \mathbf{x}_i), \\
p_{\theta}(\mathbf{x}_{i-1} \mid \mathbf{x}_i):= & \mathcal{N}(\mathbf{x}_{i-1} ; \boldsymbol{\mu}_{\theta}(\mathbf{x}_i, i), \mathbf{\Sigma}_{\theta}(\mathbf{x}_i,i)),
\end{align}
where the mean $\boldsymbol{\mu}_{\theta}$ and covariance $\mathbf{\Sigma}_{\theta}$ matrices are learned and parameterized by $\theta$ at step $i$.
The purpose of these transitions is to gradually remove the noise and slowly add structure to the samples.

The second part is the \textit{forward process}, also named the \textit{diffusion process}. It is this part that distinguishes diffusion models from other types of latent variable models. This process is not learned but fixed. Indeed, the approximate posterior $p(\mathbf{x}_{1:T} \mid \mathbf{x}_{0})$, noted $q(\mathbf{x}_{1:T} \mid \mathbf{x}_{0})$ in the following, is fixed to a Markov chain that gradually adds Gaussian noise to the data according to a variance schedule $\beta_1, \beta_2,...,\beta_T$
\begin{align}
q(\mathbf{x}_{1:T} \mid \mathbf{x}_{0}):= & \prod_{i=1}^T q(\mathbf{x}_i \mid \mathbf{x}_{i-1}), \\
q(\mathbf{x}_i \mid \mathbf{x}_{i-1}):= & \mathcal{N}(\mathbf{x}_i ; \sqrt{1-\beta_i} \mathbf{x}_{i-1}, \beta_i \mathbf{I}).
\end{align}
Finally, training is achieved by maximizing the log-likelihood. However, it is intractable as it requires marginalizing over all possible realization of the latent variables. 
Thus, it is performed by optimizing the usual variational bound on negative log-likelihood, similar to VAEs. Then, by using the fact that the noise schedule $\beta_i$ is known, it is possible to compute the posterior in closed form, which allows for efficient loss computation. 
The mathematical details of the loss, the re-parametrization trick, the forward and reverse neural network architectures implemented, and the training and sampling procedures, are provided in \cite{capeldenoising}.

\section{Assessment framework}\label{sec:quality_assessment}

The evaluation of scenarios is conducted in terms of value and quality \cite{morales2013integrating}. Quality assessment consists of using metrics to evaluate the ability of the forecasts to simulate the characteristics of the processes involved. Value assessment aims to quantify the benefits of using forecasts in decision-making applications, such as the optimal strategy for an energy retailer bidding in the electricity market. 

\subsection{Quality evaluation}\label{sec:quality_methodology}

Assessing and comparing generative models is a challenging task. There is yet to be a consensus or standard guidelines regarding which metrics are ideal for assessing the model's abilities and limitations. A model can achieve impressive results for a given metric, but it is not necessarily the case for other criteria. Thus, combining several complementary metrics to assess models is a good practice. 

This study uses five complementary quality metrics divided into two groups: (1) the \textit{univariate} metrics comprise the continuous ranked probability score (CRPS), the quantile score (QS), and the reliability diagram. They can only assess the quality of the scenarios to their marginals; (2) the \textit{multivariate} metrics are composed of the energy score (ES) and the variogram score (VS). They can directly assess multivariate scenarios.
The mathematical definitions and details of implementing these metrics are presented in \cite{DUMAS2022117871}.

The CRPS generalizes the mean absolute error for deterministic forecasts to the case of probabilistic forecasts and is one of the most widely used accuracy metrics \cite{gneiting2007strictly}. It is negatively oriented, \textit{i.e.}, the lower, the better. The CRPS is used, in our case, to assess the skill of each marginal of the forecasts, \textit{i.e.}, the twenty-four time periods of the day, of the PV, wind power, and load scenarios. 
Unlike the quantile score, the CPRS does not concentrate on any specific point of the probability distribution but considers the distribution of the forecasts as a whole.
% estimator of the CRPS NRG
The CRPS is computed over the marginals of $\mathbf{\hat{x}}$ by using the estimator of the energy form of the CRPS provided by \cite{zamo2018estimation} for a given day $d$ of the testing set.
Then, it is averaged over the entire testing set and all marginals (24 periods of the day).

The QS, known as the pinball loss score, provides details about the forecast quality at specific quantiles, \textit{i.e.}, over-forecasting or under-forecasting, particularly those related to the tails of the distribution considered \cite{lauret2019verification}. It assigns asymmetric weights to negative and positive errors for each quantile. The lower the QS, the more accurate the quantile forecast. Thus, it complements the CRPS. 
For a given day $d$ of the testing set, the QS is computed for 99 quantiles (1, 2, ..., 99-th quantile).
Then, it is averaged over 24 periods, the entire testing set, and all quantiles.

% ES -> multivariate scoring rule
The CRPS forms a particular case of a very general type of scoring rule, the ES, which is a multivariate generalization formulated and introduced by \cite{gneiting2007strictly}. The ES is the most commonly used metric to assess a finite number of scenarios modeling a distribution. Like CRPS and QS, the ES is proper and negatively oriented, \textit{i.e.}, a lower score represents a better forecast.
It evaluates forecasts relying on marginals with correct variances but biased means. 
%
% estimator of the ES
In this study, the ES is computed by the estimator provided by \cite{gneiting2008assessing} for each day of the testing test and is averaged.

% VS
However, the ES needs to be improved in detecting erroneously specified correlations between the marginals of the multivariate quantity. Therefore, the study \cite{scheuerer2015variogram} proposed the VS, which captures correlations between multivariate components based on the geostatistical concept of variograms. Similarly to the ES, it is computed for each day of the testing set and averaged.

\subsection{Value evaluation}\label{sec:forecast_value}

Similarly to \cite{DUMAS2022117871}, the forecast value is assessed by considering the case study of an electricity retailer bidding on the day-ahead market. The retailer aims to balance its production and consumption portfolio hourly to avoid financial penalties in case of imbalance by trading the surplus or deficit of energy. In this case study, we include a battery energy storage system. Thus, it allows for managing the portfolio optimally by minimizing imports when day-ahead prices are prohibitive and maximizing exports when prices are advantageous.

% Scenario-based formulation
A two-stage stochastic optimization formulation, detailed in \cite{DUMAS2022117871}, with linear constraints models this case study. It uses a scenario approach where load, PV, and wind power uncertainties are modeled with trajectories of these variables computed by the generative models. The first-stage variables are the day-ahead bids of the electricity retailers, which do not depend on uncertainties. The second stage variables, scenario dependent, are the dispatch decisions in wind power and PV production, where curtailment is allowed, and the charge or discharge of the battery. A cost function models the imbalance penalties. Thus, the planner aims to minimize the imbalances and maximize the profits on the day-ahead market.

\section{Numerical Results}\label{sec:numerical_results}

\begin{figure}[tb]
\centering
\includegraphics[width=\linewidth]{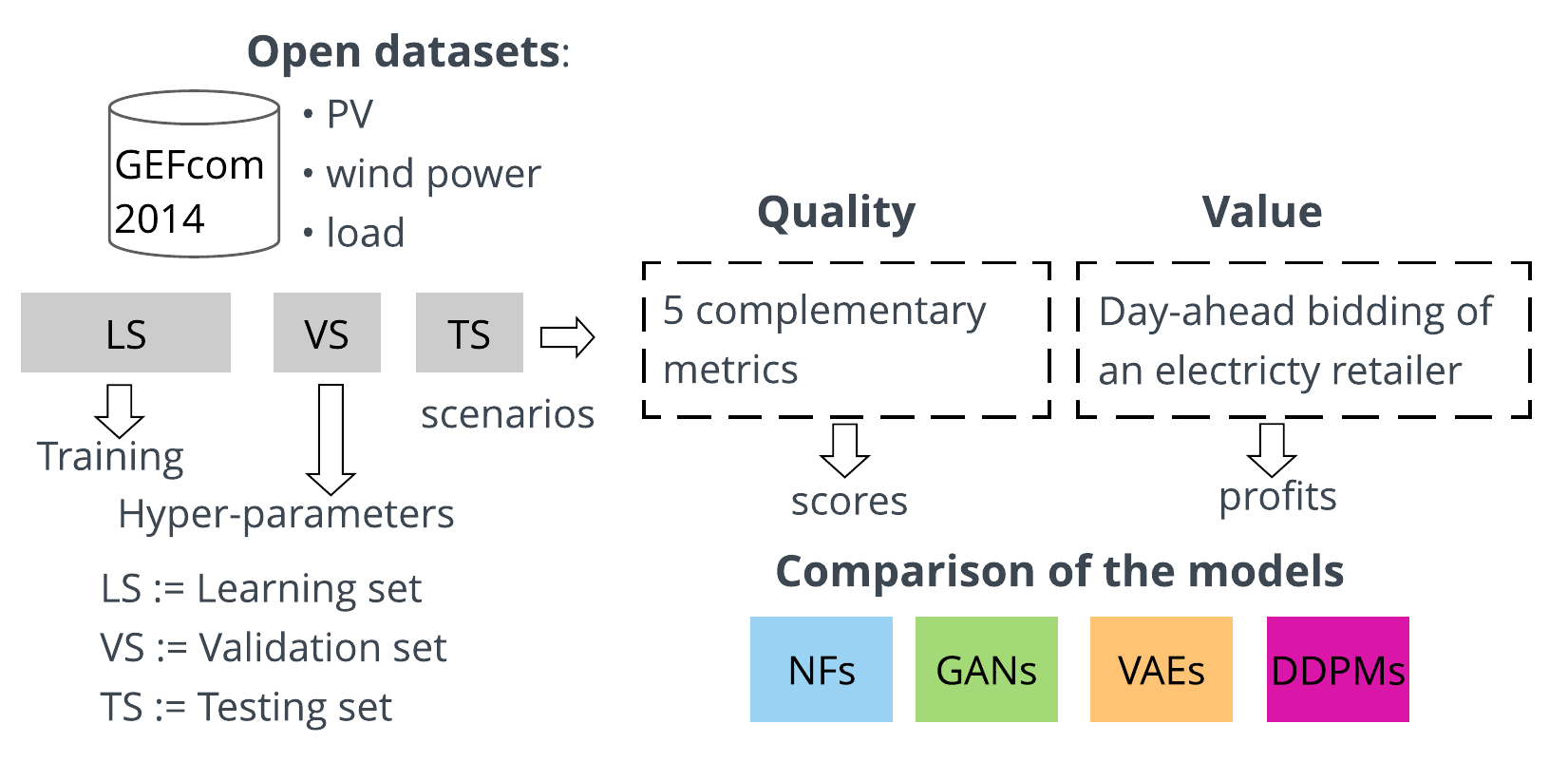}
\caption{Approach adopted to assess the quality and value of the models implemented: GAN, VAE, NF, and DDPM.}
\label{fig:numerical-experiments-methodology}
\end{figure}

The open-access GEFCom 2014 dataset \cite{hong2016bprobabilistic} comprises one, ten, and three zones for load, wind, and PV tracks, respectively. The quality and value assessments of the generative models are performed on this easily reproducible case study.
Fig.~\ref{fig:numerical-experiments-methodology} depicts the overall approach adopted to assess the quality and value of the models implemented.
The PV, wind power, and load datasets are randomly divided into learning, validation, and testing sets. The models are trained on the learning set, the validation set is used to select the optimal hyper-parameters of each model, and the testing set allows the quality and value assessment. 

\subsection{Quality results}\label{sec:quality_res}

\begin{figure}[tb]
\centering
\begin{subfigure}{0.25\textwidth}
\centering
\includegraphics[width=\textwidth]{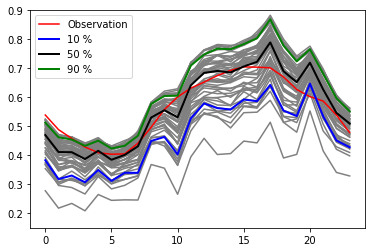}
\caption{GAN}
        \end{subfigure}%
 \begin{subfigure}{0.25\textwidth}
 \centering
\includegraphics[width=\textwidth]{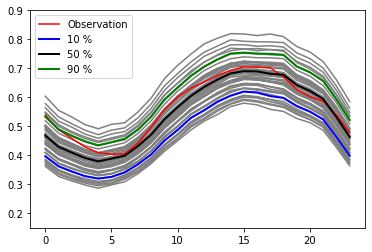}
     \caption{VAE}
        \end{subfigure}
\begin{subfigure}{0.25\textwidth}
            \centering
\includegraphics[width=\textwidth]{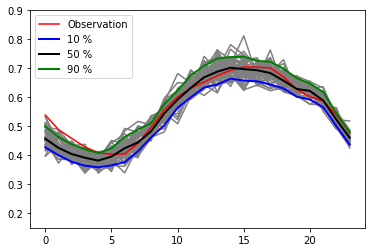}
            \caption{NF}
\end{subfigure}%    
\begin{subfigure}{0.25\textwidth}
            \centering
\includegraphics[width=\textwidth]{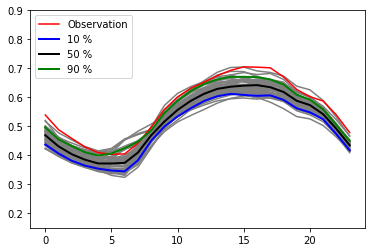}
            \caption{DDPM}
        \end{subfigure}
\caption{Load scenarios shape comparison.}
        \label{res:load}
\end{figure}

\begin{figure}[tb]
\centering
\begin{subfigure}{0.25\textwidth}
\centering
\includegraphics[width=\textwidth]{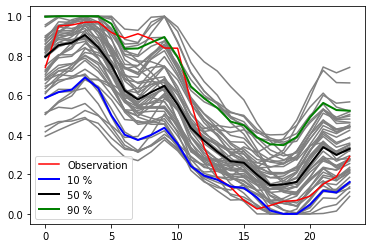}
\caption{GAN}
        \end{subfigure}%
 \begin{subfigure}{0.25\textwidth}
 \centering
\includegraphics[width=\textwidth]{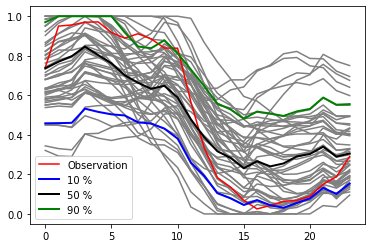}
     \caption{VAE}
        \end{subfigure}
\begin{subfigure}{0.25\textwidth}
            \centering
\includegraphics[width=\textwidth]{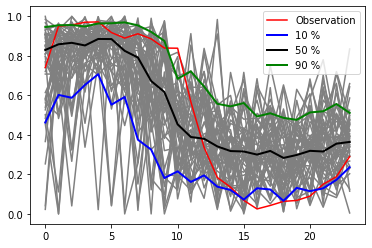}
            \caption{NF}
\end{subfigure}%    
\begin{subfigure}{0.25\textwidth}
            \centering
\includegraphics[width=\textwidth]{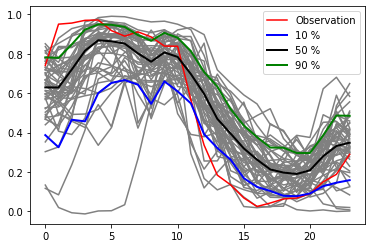}
            \caption{DDPM}
        \end{subfigure}
\caption{Wind scenarios shape comparison.}
        \label{res:wind}
\end{figure}

\begin{figure}[tb]
\centering
\begin{subfigure}{0.25\textwidth}
\centering
\includegraphics[width=\textwidth]{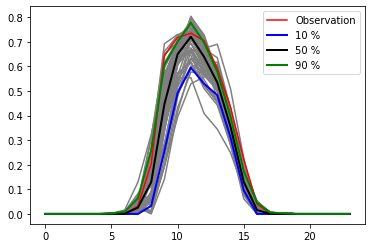}
\caption{GAN}
        \end{subfigure}%
 \begin{subfigure}{0.25\textwidth}
 \centering
\includegraphics[width=\textwidth]{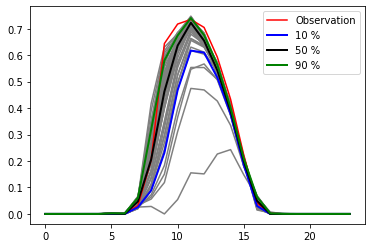}
     \caption{VAE}
        \end{subfigure}
\begin{subfigure}{0.25\textwidth}
            \centering
\includegraphics[width=\textwidth]{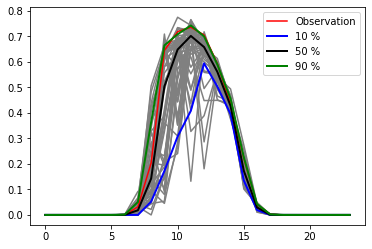}
            \caption{NF}
\end{subfigure}%    
\begin{subfigure}{0.25\textwidth}
            \centering
\includegraphics[width=\textwidth]{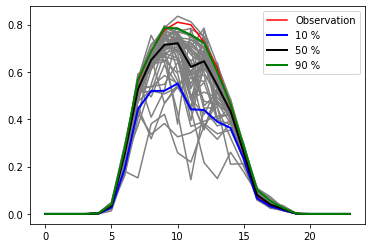}
            \caption{DDPM}
        \end{subfigure}
\caption{PV scenarios shape comparison.}
        \label{res:pv}
\end{figure}

Each model generates one hundred scenarios per day of the testing set. They are used to compute the selected metrics to perform the quality assessment.
Table~\ref{tab:quality_average_scores} presents the quality scores averaged over the testing set. Notice that the MAE-r is the mean absolute error between the reliability curve and the diagonal.
\begin{table}[tb]
\renewcommand{\arraystretch}{1.25}
\begin{center}
\caption{Averaged quality scores over the testing set per dataset. The best-performing model for each track is written in bold. 
The CRPS, QS, MAE-r, and ES are expressed in \%.}
\begin{tabular}{clrrrr}
\hline  \hline
\multicolumn{1}{l}{}  &  & VAE            & GAN   & NF   & DDPM           \\ \hline
% wind
\multirow{5}{*}{Wind} & ES    & 54.82 & 60.52 & 56.71          & \textbf{54.47}          \\
& VS    & 17.87          & 19.87 & 18.54          & \textbf{17.29} \\
& QS    & 4.45  & 4.95  & 4.58           & \textbf{4.41}           \\
& CRPS  & 8.80  & 9.79  & 9.07           & \textbf{8.73}         \\
& MAE-r & 2.67           & 6.82  & 2.83           & \textbf{1.35}  \\ \hline
% PV
\multirow{5}{*}{PV}   & ES    & 24.65          & 24.15 & 23.08 & \textbf{21.60}          \\
& VS    & 5.02           & 4.88  & 4.68  & \textbf{4.16}           \\
& QS    & 1.31           & 1.32  & 1.19  & \textbf{1.14 }         \\
& CRPS  & 2.60           & 2.61  & 2.35  & \textbf{2.26}           \\
& MAE-r & 9.04           & 4.94  & \textbf{2.66}           & 8.06  \\ \hline
% Load
\multirow{5}{*}{Load} & ES    & 15.11          & 17.96 & \textbf{9.17}  & 9.76         \\
& VS    & 1.66           & 3.81  & 1.63  & \textbf{1.49}          \\
& QS    & 1.39           & 1.52  & \textbf{0.76}  & 0.8          \\
& CRPS  & 2.74           & 3.01  & \textbf{1.51}  & 1.69          \\
& MAE-r & 13.97          & 9.99  & \textbf{7.70}          & 9.43 \\ \hline  \hline
\end{tabular}
\label{tab:quality_average_scores}
\end{center}
\end{table}
DDPM outperforms VAE, GAN, and NF models for all the quality metrics on the wind and PV tracks, except for the MAE-r on the PV track. On the load track, the NF model achieves the best scores except for the VS, where DDPM is the best. 
Fig.~\ref{res:load}, \ref{res:wind}, and \ref{res:pv} depict for each model 50 scenarios generated (grey) for a given day selected randomly from the testing set, along with the ten \% (blue), 50 \% (black), 90 \% (green) quantiles, and the observations (red), of the load, wind, and PV tracks.
The shape of the NF scenarios differs significantly from the GAN and VAE as they tend to be more variable with no identifiable trend. In contrast, the VAE and GAN scenarios vary mainly in nominal power but have similar shapes. The DDPM scenarios have an identifiable trend but seem less correlated than the scenarios of the GAN and VAE models. However, they tend to underestimate the load and PV power compared to the other models, which further investigations must address.

\subsection{Value results}\label{sec:value_res}

% Dataset description
The electricity retailer portfolio comprises wind power, PV generation, load, and a battery energy storage device. One thousand five hundred independent simulated days are considered by combining the 50 days of the testing set with the 30 possible PV and wind generation zones (three solar and ten wind farms).
%
% Strategy to compute the profit
A two-step approach assesses the value of the four generative models: DDPM, NF, GAN, and VAE.
First, for each generative model and the 1 500 days simulated, the two-stage stochastic planner computes the day-ahead bids using the PV, wind power, and load scenarios. 
Then, a real-time dispatch uses the PV, wind power, and load observations, with the day-ahead bids previously computed as parameters. 
%
% Profit computation description
The net profit is the gross profit, the electricity sold on the day-ahead market minus the purchases, minus the imbalance penalties. It is computed for each simulation day (1 500) and summed in Table~\ref{tab:value-res}. The DDPM achieves the highest net profit, followed by the NF. However, there is room for improvement. An oracle, which has perfect future knowledge, achieves 300~k\euro.
\begin{table}[tb]
\renewcommand{\arraystretch}{1.25}
\centering
\caption{Total net profit (k\euro) per model.}
\begin{tabular}{lrrrr}
\hline \hline
& GAN & VAE & NF  & DDPM  \\ \hline
Net profit (k\euro) & 93  & 97  & 107 & \textbf{112}   \\ \hline \hline
\end{tabular}
\label{tab:value-res}
\end{table}

\subsection{Synthesis of results}\label{sec:discussion}

Table \ref{tab:comparison} outlines the key results of this study by comparing the models via readily star ratings: the more stars, the better. In addition to the value and quality assessments, it suggests other comparison criteria such as the training and sample computation times, the hyper-parameter search and sensibility, and the ease of implementation.
Specifically, for each dataset, training and sample computation times are evaluated on the total reported training and generating times. The hyper-parameters search is assessed by the number of configurations tested before reaching satisfactory and stable results over the validation set. The deviations in the quality metrics from the optimal hyper-parameter values evaluate the sensitivity of a given model to its hyper-parameters. Finally, the implementation-friendly criterion is appraised regarding the mathematical complexity of the model and the amount of knowledge required to implement it.
\begin{table}[tb]
\renewcommand{\arraystretch}{1.25}
\begin{center}
\caption{Comparison of the deep generative models.
}
\begin{tabular}{lcccc}
\hline \hline
Criteria			     & VAE          &  GAN         & NF    & DDPM        \\ \hline
Train speed              & $\nstars[4]$ & $\nstars[3]$ & $\nstars[2]$   & $\nstars[1]$ \\
Sample speed             & $\nstars[4]$ & $\nstars[3]$ &  $\nstars[2]$  & $\nstars[1]$ \\
Quality                  & $\nstars[2]$ & $\nstars[1]$ & $\nstars[3]$  &  $\nstars[4]$\\
Value                    & $\nstars[2]$ & $\nstars[1]$ & $\nstars[3]$  & $\nstars[4]$ \\
Hp search                & $\nstars[3]$ & $\nstars[1]$ & $\nstars[4]$ & $\nstars[2]$\\
Hp sensibility 	         & $\nstars[3]$ & $\nstars[1]$ & $\nstars[4]$ & $\nstars[2]$\\
Implementation           & $\nstars[4]$ & $\nstars[3]$ & $\nstars[1]$ & $\nstars[2]$\\
\hline \hline
\end{tabular}
\label{tab:comparison}
\end{center}
\end{table}
The VAE is the fastest model for training and generating scenarios, followed by the GAN, NF, and DDPM models. DDPM is particularly slow to train and generate samples due to its sequential nature. Furthermore, DDPM and NF have the drawback of having latent spaces of the same dimension as the input dimension leading to expensive computations.
Concerning the hyper-parameters search and sensibility, the NF model is the most accessible to calibrate. Compared with the VAE, DDPM, and GAN, we found relevant hyper-parameter values by testing only a few combinations. The NF is the most robust to hyper-parameter modifications, followed by the VAE and DDPM models. The GAN is the most sensitive, with variations of the hyper-parameters that may result in very poor scenarios in terms of quality and shape. 
In this study, the VAE is an effortless model to implement, as the encoder and decoder are uncomplicated feed-forward neural networks. The most significant difficulty lies in the reparameterization trick. The gradient penalty of the GAN considered makes it more challenging to implement. However, the neural network architectures of the discriminator and the generator do not contain any difficulties, as they are feed-forward neural networks similar to the VAE. The NF and DDPM are the most challenging models to implement. 
However, forecasting practitioners do not necessarily have to implement generative models and can use numerous existing Python libraries. 

\section{Conclusion}\label{sec:conclusion}

This study uses the open data of the Global Energy Forecasting Competition 2014 to assess the quality and value of the denoising diffusion probabilistic model with state-of-the-art deep learning generative models: normalizing flows, generative adversarial networks, and variational autoencoders. 
The models employ weather forecasts to generate improved PV, wind, and load scenarios. 
The results demonstrate that denoising diffusion probabilistic models challenge the other generative models with better quality scores and the highest profits regarding the value of the electricity retailer case study.

In future work, four limitations could be addressed. First, in the current study, the variance of the reverse process is set to a fixed constant. However, the studies \cite{IDDPM} and \cite{diff_beat_gan} have demonstrated improvements in sample quality in computer vision applications by learning this parameter.
Second, the diffusion steps contribute differently to the quality of the samples. The paper \cite{IDDPM} proposes to adjust the noise schedule to balance the relative importance of each diffusion step through the entire diffusion process.
Third, the papers \cite{IDDPM} and \cite{diff_wave} propose to use fewer sampling diffusion steps than those employed in training to decrease the training and sampling computation time.
Finally, improvements to the current architecture could be made, particularly the conditioner, a simple multi-layer perceptron. For instance, a recurrent neural network could be implemented to leverage the sequential nature of the condition vector. 

\section*{Acknowledgment}

The authors would like to thank professor Bertrand Corn\'elusse of Liege University, as this work was conducted in its team, and Emily Little of RTE R\&D, that helped to review the paper.

\bibliographystyle{ieeetr}
\bibliography{biblio}

\end{document}